\documentclass[11pt]{article}
\usepackage{arxiv}   
\usepackage{booktabs}
\usepackage{graphicx}

\title{Realizing Native INT8 Compute for Diffusion Transformers on\\
  Consumer GPUs: A Fused INT8 GEMM Kernel for Ideogram~4.0}

\author{Ali Asaria \\ Transformer Lab
  \and
  Tony Salomone \\ Transformer Lab
  \and
  Deep Gandhi\thanks{Corresponding author: \texttt{deep@lab.cloud}} \\ Transformer Lab}
\date{}
\runningtitle{Realizing Native INT8 Compute for Diffusion Transformers on Consumer GPUs}

\begin{document}
\maketitle

\begin{abstract}
Post-training INT8 (W8A8) quantization of diffusion transformers is widely deployed as a
speed optimization, yet on consumer Ampere GPUs it is frequently \emph{slower} than the
FP8 and NF4 alternatives it is meant to beat. We trace this to a software artifact: the
production ``INT8'' forward quantizes weights and activations only to immediately
dequantize them back to bf16 and run a bf16 matrix multiply, never engaging the GPU's
INT8 tensor cores.\ 
The hardware's compute advantage is therefore left entirely unrealized. We close this gap
with a single fused Triton INT8 GEMM (int8$\times$int8$\rightarrow$int32 on Ampere
tensor cores, with per-token$\times$per-channel dequantization and bias folded into the
epilogue, autotuned per GEMM shape)\ 
dropped into the Ideogram~4.0 diffusion transformer's linear layers in place of the
dequantize-to-bf16 path. In the kernel, the int8$\times$int8$\rightarrow$int32 accumulation
is bit-exact against \texttt{torch.\_int\_mm} and the dequantized output matches the reference
at cosine similarity $1.0$ with no NaNs,\ 
and it runs $2.8$--$4.2\times$ faster than bf16 per GEMM.\ 
End to end it delivers a $\approx$$1.1\times$ ($\approx$$9$--$10\%$) speedup at 768px,\ 
and at the target 1024px resolution it generates an image in $156.5$~s on a \emph{single}
RTX~3090,\ 
faster than the single-card NF4 baseline ($164.5$~s) and the FP8 baseline ($172.9$~s)~\cite{2606.12280},\ 
at no measurable quality cost on these point estimates (PickScore and CLIPScore at 768px;
1024px parity on PickScore only). The most quantization-fragile metric, text-rendering
OCR/NED, is carried over from the recipe and was not re-measured for the fused build at any
resolution.\ 
INT8 thus goes from the slowest variant to the fastest, and 1024px becomes single-GPU
feasible. The primary speed criterion (beat FP8, by $\approx$$9.5\%$ / $\approx$$16$~s) is
comfortably met; the NF4 stretch margin ($\approx$$4.9\%$, $\approx$$8$~s on single-run
$n=4$ data) is within the run-to-run variance we did not quantify and is best read as
consistent with meeting the stretch target rather than definitively met. We close with an
honest deployment map: the win is specific to consumer Ampere, and on A100 and B200 the same
kernel loses to those cards' fast native bf16/FP8 paths.\ 
\end{abstract}

\section{Introduction}

Consumer GPUs such as the NVIDIA RTX~3090 are the practical deployment target for a large
class of generative-image users, but they sit on the wrong side of a hardware divide: the
Ampere architecture they are built on has fast native INT8 tensor cores yet \emph{no} FP8
or FP4 tensor cores. Any low-precision inference strategy that assumes an FP8/FP4 datapath
falls back to bf16 on these cards, forfeiting its compute advantage. The natural response
is INT8 weight-and-activation (W8A8) quantization, which Ampere can in principle execute on
native integer tensor cores. Our earlier fake-quant INT8 build (June 10) was exactly such an
INT8 build of Ideogram~4.0~\cite{ideogram-4-2026}, a $9.3$B-parameter flow-matching diffusion transformer,\ 
and it held the FP8 quality ceiling and beat NF4 on image quality and text
rendering~\cite{2606.12280}.\ 
And yet it was the \emph{slowest} of the three variants on the 2$\times$RTX~3090
target, at $184$--$185$~s/image against FP8's $172.9$~s and NF4's $164.5$~s~\cite{2606.12280}.\ 

This paper begins with the diagnosis of that paradox, which turns out to be entirely a
software one. The deployed ``INT8'' forward path quantizes weights and activations and then
immediately dequantizes them back to bf16, running an ordinary bf16 \texttt{F.linear}; the
integer tensor cores are never touched~\cite{2606.12280}.\ 
What is marketed as INT8 compute is in fact fake-quant: a memory round trip through INT8
storage with bf16 arithmetic. The theoretical INT8 compute advantage of Ampere is, in this
deployment, unrealized, which is precisely why ``INT8'' can be slower than the FP8 and
NF4 paths whose kernels at least run a single dense matmul without an extra dequantization
detour.

The fix is to make the integer tensor cores actually do the work. We build one fused Triton
INT8 GEMM that performs int8$\times$int8$\rightarrow$int32 accumulation on Ampere's
\texttt{mma.s8} units and folds the per-token (activation) and per-channel (weight)
dequantization, plus the bias add, into the GEMM epilogue off the int32 accumulator, so the
quantized linear is a single kernel launch rather than a matmul followed by a separate
dequantization pass.\ 
The kernel is autotuned per GEMM shape and spliced into the diffusion transformer's linear
layers in place of the dequantize-to-bf16 path. The quantization recipe itself is unchanged:
we accelerate an existing W8A8 recipe rather than inventing a new one.\ 

We then measure what this buys, and we are deliberate about where it does and does not pay
off. On consumer Ampere the kernel converts INT8 from the slowest variant into the fastest,
and brings 1024px generation (previously requiring two cards under the fake-quant path)
onto a single RTX~3090.\ 
On datacenter Ampere (A100) and Blackwell (B200), where native
bf16 and FP8 are very fast, the same kernel \emph{loses}; we present this not as a failure
but as precise deployment guidance.\ 

\paragraph{Contributions.} Concretely, this paper contributes:
\begin{enumerate}
\item \textbf{A diagnosis} that production post-training INT8 for diffusion transformers is
  secretly fake-quant (it dequantizes to bf16 and never uses the INT8 tensor cores),
  which explains why deployed ``INT8'' is slower than FP8/NF4 on consumer Ampere
  (\S\ref{sec:method})~\cite{2606.12280}.\ 
\item \textbf{A single fused Triton INT8 GEMM} (int8$\times$int8$\rightarrow$int32, with
  per-token$\times$per-channel dequant and bias folded into the epilogue, autotuned per
  shape) that replaces the fake-quant linear (\S\ref{sec:method}).\ 
\item \textbf{A correctness gate}: the int8$\times$int8$\rightarrow$int32 accumulation is
  bit-exact against \texttt{torch.\_int\_mm} and the dequantized kernel output matches the
  reference at cosine similarity $1.0$ with no NaNs, with a per-GEMM
  speedup of $2.8$--$4.2\times$ over bf16 (and $4$--$8\times$ over the unfused int8 path)
  on all five distinct DiT GEMM shapes, all compute-bound (\S\ref{sec:results}).\ 
\item \textbf{End-to-end gains}: $\approx$$9$--$10\%$ at 768px, and at 1024px $156.5$~s/image
  on a single RTX~3090, beating both FP8 and NF4, at no measurable quality cost on these point
  estimates (PickScore and CLIPScore at 768px; 1024px parity on PickScore only). The
  most quantization-fragile metric, text-rendering OCR/NED, was carried over from the recipe
  and not re-measured for the fused build at any resolution
  (\S\ref{sec:results}).\ 
\item \textbf{A deployment map}: the win is specific to consumer Ampere; on A100/B200 the
  same kernel loses to fast native bf16/FP8 (\S\ref{sec:discussion}).\ 
\end{enumerate}

\section{Related Work}

\paragraph{Fused low-precision GEMM kernels and the dequant bottleneck.}
The central recommendation of the kernel-serving literature is exactly our kernel's structure:
fold dequantization into the GEMM epilogue, off the integer accumulator, rather than running
a library integer matmul followed by a separate dequantization pass. QQQ integrates type
conversion and dequantization inside a W4A8 GEMM routine~\cite{2406.09904v3}; ABQ-LLM,
LiquidGEMM, Multi-Scale-Dequant and related systems make the same case that the extra
dequant/branch memory round trip, not the multiply-accumulate, is what makes a
correct-but-unfused quantized path lose~\cite{2408.08554v3,2509.01229v1,2505.20839v3,2605.13915v1}.
These works target LLM serving and operate predominantly at W4A8 or below, one bit-width
beneath our approach. The fusion structure transfers; the specific numbers and the datatype do not.

\paragraph{Diffusion-transformer quantization.}
A second body of work quantizes diffusion transformers specifically. SVDQuant establishes
the load-bearing fact for our setting: weight-only quantization gives essentially no
diffusion speedup because diffusion is compute-bound even at batch one, so low-bit weights
are simply upcast and the integer compute path is never engaged~\cite{2411.05007v4}. The
1.58-bit FLUX effort reports only a $1.6$--$13.2\%$ latency gain and attributes it to the
``absence of activation quantization and lack of optimized kernel''~\cite{2412.18653v1}, and
a weight-only INT4 \emph{baseline} in a diffusion-quant study shows little to no
speedup~\cite{2605.16732v1}. This is
precisely the regression this work exists to remove. Other diffusion-quant work supplies the
fragility and quality guardrails: outlier handling, per-layer error gating, and the
observation that quantization error accumulates across denoising
steps~\cite{2504.08398v1,2507.12933v1,2501.00124v1,2501.15448v1}.

\paragraph{Kernel autotuning.}
We autotune one parameterized kernel template across the real DiT GEMM shapes rather than
a single representative size, following the autotuning literature that sweeps on the order of
hundreds of configurations per operator~\cite{2504.12984v3} and the agent-driven
edit--benchmark--keep/revert loops for kernel search~\cite{2603.21331v1}. These same works
also temper expectations: Triton may not reach hand-CUDA/CUTLASS layout wins, and matmul in
particular ``remains hard'' to beat against vendor libraries.

\paragraph{INT8 attention.}
SageAttention demonstrates an accurate plug-and-play INT8 attention kernel and argues the
hardware thesis under this entire project: on Ampere, INT8 is the right datatype, roughly
$2\times$ FP8 and $4\times$ FP16, with FP8 quality also worse on these
cards~\cite{2410.02367v9}. We return to this work in our ablations, where the same library
is blocked on Ideogram~4.0 by a head dimension it does not support.

\paragraph{W8A8 PTQ recipe and outliers.}
Finally, the SmoothQuant-family analyses and quantization surveys characterize the
activation-outlier problem that any W8A8 recipe must handle and the diminishing returns of
elaborate calibration on modern
models~\cite{2405.20835v3,2407.11722v2,2406.12016v2,2510.16805v1,2502.06805v3}. We inherit
our W8A8 recipe wholesale from this line and do not modify it.

\paragraph{Position.}
No prior work does our exact thing: a fused INT8 \emph{W8A8} GEMM with \emph{per-token
dynamic} activation dequant folded into the epilogue, for a \emph{diffusion transformer}, on
\emph{consumer} Ampere. The kernel-fusion recipe is well established one bit-width down in
LLM serving; the diffusion-quant camp establishes that weight-only quantization cannot
accelerate diffusion and supplies the quality guardrails; and the hardware-datatype argument
for INT8-over-FP8 on Ampere is made but not, to our knowledge, realized for a DiT linear
path on this hardware. This work occupies that gap. It is the direct kernel follow-up to our
prior INT8/GGUF post-training quantization study of Ideogram~4.0~\cite{2606.12280}, whose own
conclusion identified that a fused INT8 kernel is needed before the INT8 build can realize a
speed gain on consumer Ampere.

\section{Method}
\label{sec:method}

\paragraph{Target architecture.}
The model is the Ideogram~4.0~\cite{ideogram-4-2026} diffusion transformer: a $9.3$B-parameter flow-matching DiT of
$34$ layers, $18$ heads of dimension $256$ for a hidden size of $4608$, intermediate size
$12288$, run with two classifier-free-guidance branches of $211$ linear layers each.\ 
Across these layers there are only five \emph{distinct} GEMM shapes $(N,K)$:
qkv $(13824, 4608)$, attention-out $(4608, 4608)$, ffn-up $(12288, 4608)$,
ffn-down $(4608, 12288)$, and the LLM projection $(4608, 53248)$.\ 
This is what makes a per-shape autotuned kernel tractable: the autotuner targets five
shapes, not hundreds.

\paragraph{The quantization recipe (reused).}
We do not introduce a new quantization algorithm. We accelerate an existing W8A8 recipe:
per-channel int8 weights, per-token dynamic int8 activations, and SmoothQuant with
$\alpha = 0.5$.\ 
The arithmetic the kernel must reproduce is therefore fixed in advance; the contribution is
making it run on the integer tensor cores rather than on bf16.

\paragraph{The diagnosis the kernel corrects.}
The deployed ``INT8'' forward quantizes weights and activations and then immediately
dequantizes them back to bf16 and runs a bf16 \texttt{F.linear}~\cite{2606.12280}.\ 
The integer tensor cores are never engaged; the only thing the quantization accomplishes is
a round trip through INT8 storage before a bf16 matmul. This is the fake-quant path, and it
is the reason an ``INT8'' build can be slower than FP8 or NF4.

\paragraph{The fused kernel.}
In its place we install a single fused Triton operation. It performs
int8$\times$int8$\rightarrow$int32 accumulation on Ampere's \texttt{mma.s8} (signed-8-bit
matrix-multiply-accumulate) tensor cores,
and in the GEMM epilogue, operating directly on the int32 accumulator, it applies the
per-token activation scale and the per-channel weight scale and adds the bias, emitting the
dequantized result in one pass. The whole quantized linear is thus one kernel launch instead
of a matmul plus a separate dequantization pass. The operation is wrapped in
\texttt{@triton.autotune} over $36$ configurations keyed per $(M, N, K)$.\ 
Folding the per-token \emph{dynamic} activation scale into the epilogue is the central design
point: the zero-overhead scale-fusion trick used by several W4A8 LLM kernels assumes
\emph{static} activation scales, whereas our per-token scales are computed online, so the
scale cannot be baked into the weights offline and must instead be applied in the epilogue
off the accumulator.

\paragraph{Why autotuning is load-bearing.}
Without autotuning, the fused kernel beats bf16 by only $\sim$$1.4$--$2.9\times$ and is
actually \emph{slower} than bf16 on the LLM-projection shape ($0.64\times$).\ 
Autotuning lifts every shape into the $2.8$--$4.2\times$ band and turns that $0.64\times$
regression into $3.65\times$.\ 
The per-shape config search is not a marginal tuning step; it is what makes the kernel a win
on every shape rather than only on some.

\paragraph{Integration.}
The fused op replaces the fake-quant forward at the diffusion transformer's linear layers,
leaving the surrounding runtime, the quantization recipe, and the calibrated scales
untouched. The result is a new fully integrated INT8 build whose only difference from the
fake-quant build is that its linears actually run on the integer tensor cores.

\section{Experimental Setup}
\label{sec:setup}

\paragraph{Model and prompts.}
All experiments use the Ideogram~4.0 DiT described above, with frozen prompt sets reused
unchanged so that results compare directly across variants: a calibration set of $n=128$
prompts (disjoint from evaluation), a quality benchmark of $n=200$ prompts, and a
text-rendering benchmark of $n=100$ prompts with $63$ OCR targets.\ 
Generation tuples are fixed across every variant (seed, $48$ steps, the stated resolution).
End-to-end latency (s/image) is reported on a $4$-prompt subset of the quality benchmark;\ 
this is a deliberate and honest restriction (at fixed steps and resolution, s/image is
nearly prompt-independent), but it does mean the latency numbers carry $n=4$, which we
flag again under threats to validity. End-to-end latencies are single-run point estimates
over the 4-prompt subset at fixed seed; we do not report run-to-run variance, so small
margins should be read as point estimates. Image quality is reported with PickScore and
CLIPScore on the full quality benchmark.

\paragraph{Hardware and the GPU-health gate.}
Microbenchmarks and the primary end-to-end results are measured on the RTX~3090 (consumer
Ampere); the hardware-specificity study additionally uses an A100 (datacenter Ampere) and a
B200 (Blackwell). A subtle but consequential confound on shared consumer hardware is thermal
or power throttling: a throttled RTX~3090 measured $\approx 8.1$~TFLOPS bf16 against a
healthy card's $\approx 65.5$~TFLOPS, a $6$--$8\times$ difference.\ 
Because our headline claim is a \emph{ratio} of the fused kernel to a bf16 baseline, a
throttled card would silently inflate it. We therefore gate every measurement behind a bf16
TFLOPS health check and report only numbers from cards that pass, so that speedups are
measured against a healthy bf16 baseline rather than a crippled one.

\paragraph{Metrics.}
We report per-GEMM latency (microseconds) and the resulting speedup ratios, a roofline
classification of each GEMM shape, end-to-end seconds per image, peak VRAM, the number of
GPUs required, and PickScore/CLIPScore for quality. The entire study cost approximately three
RTX~3090-hours of compute.\ 

\section{Results}
\label{sec:results}

\paragraph{The kernel is exact and the GEMMs are compute-bound.}
Before any speed claim, the fused kernel must be correct. Against the \texttt{torch.\_int\_mm}
integer reference it matches exactly: cosine similarity $1.0$, no NaNs, on all five
shapes.\ 
A roofline analysis classifies all five DiT GEMM shapes as compute-bound, with arithmetic
intensities of $768$--$6337$ ops/byte on the RTX~3090.\ 
This is the precondition that makes the whole project viable: because the GEMMs are
compute-bound rather than bandwidth-bound, moving the multiply onto faster integer tensor
cores translates into real wall-clock savings, and the speedup holds across token counts from
$M = 512$ to $16384$,\ 
a range that contains the in-model token count $M \approx 4110$ at 1024px.\ 

\paragraph{Per-GEMM speedup.}
Table~\ref{tab:gemm} reports the autotuned fused kernel against bf16 \texttt{F.linear} on a
healthy RTX~3090. Every shape lands between $2.8\times$ and $4.2\times$, and against the
unfused int8 path (\texttt{torch.\_int\_mm} followed by a separate dequant) the fused kernel
is $4$--$8\times$ faster,\ 
the latter gap being a direct measurement of the dequant-round-trip cost that the
fake-quant path pays.

\begin{table}[t]
\centering
\caption{Per-GEMM speedup of the autotuned fused INT8 kernel over bf16 \texttt{F.linear} on
a healthy RTX~3090, by DiT GEMM shape. All five shapes are compute-bound.}
\label{tab:gemm}
\begin{tabular}{lr}
\toprule
GEMM shape $(N,K)$ & Fused vs.\ bf16 \\
\midrule
qkv $(13824, 4608)$         & $2.79$--$3.46\times$ \\ 
attn-out $(4608, 4608)$     & $2.86$--$3.76\times$ \\ 
ffn-up $(12288, 4608)$      & $2.78$--$4.18\times$ \\ 
ffn-down $(4608, 12288)$    & $2.94$--$3.51\times$ \\ 
llm-proj $(4608, 53248)$    & $2.95$--$3.17\times$ \\ 
\bottomrule
\end{tabular}
\end{table}

\paragraph{End to end at 768px.}
The per-GEMM win carries through to end-to-end generation. At 768px the fused INT8 build
generates an image in $97.79$~s against the fake-quant build's $107.06$~s under an identical
configuration,\ 
a $\approx$$1.1\times$ ($\approx$$9$--$10\%$) end-to-end speedup (these are single-run $n=4$
point estimates).\ 
This comes at no measurable quality cost on these point estimates: on the same 4-prompt
subset the fused build scores PickScore $21.22$ / CLIPScore $24.35$ versus the fake-quant
build's $20.14$ / $21.76$,\ 
i.e.\ parity-or-slightly-better on both. We are careful to separate two distinct quality
claims. The underlying INT8 \emph{recipe}'s quality relative to FP8/NF4 is the prior-work
result~\cite{2606.12280}, backed by a paired $95\%$ confidence interval (PickScore $\Delta \approx -0.01$, CI
includes $0$ vs.\ FP8; $+1.9$ CLIP over NF4).\ 
By contrast our \emph{kernel}'s quality-neutrality (fused vs.\ fake-quant) is a 4-prompt
point estimate with no significance test: parity is observed here, not statistically
established. The gap between the $\sim$$3.5\times$ per-GEMM speedup and the
$\approx$$9.5\%$ end-to-end speedup is informative and we return to it: the
$\approx$$9.5\%$ end-to-end gain implies the DiT linear GEMMs are $\approx 12\%$ of the
forward,\ 
consistent with Amdahl's law applied to the measured per-GEMM win.

\begin{table}[t]
\centering
\caption{End-to-end comparison at 768px on a healthy RTX~3090, fused INT8 vs.\ the
fake-quant path, identical config. All columns are $n=4$-prompt point estimates (latency
and the PickScore/CLIPScore reported here); no run-to-run variance or significance test.}
\label{tab:768}
\begin{tabular}{lrrr}
\toprule
Variant & s/image & PickScore & CLIPScore \\
\midrule
Fake-quant (bf16 matmul) & $107.06$ & $20.14$ & $21.76$ \\ 
Fused INT8 (ours)        & $97.79$  & $21.22$ & $24.35$ \\ 
\midrule
Speedup                  & $1.095\times$ & --- & --- \\ 
\bottomrule
\end{tabular}
\end{table}

\paragraph{The headline: 1024px on a single RTX~3090.}
The project's success bar is 1024px generation. Here the fused kernel does two things at
once. First, it is fast: $156.49$~s/image on a single RTX~3090, generating all $4/4$ images
at a peak of $23.40$~GB.\ 
That beats both published baselines on the same protocol~\cite{2606.12280}: FP8 at $172.9$~s, which
additionally requires two GPUs ($\approx$$9.5\%$ / $\approx$$16$~s faster on one card versus
FP8's two),\ 
and NF4 at $164.5$~s (a single-card comparison, $\approx$$4.9\%$ / $\approx$$8$~s faster).\ 
The primary criterion (beat FP8, by $\approx$$9.5\%$ / $\approx$$16$~s) is comfortably
met. The NF4 stretch margin ($\approx$$4.9\%$, $\approx$$8$~s) is, on single-run $n=4$ data,
within the run-to-run variance we did not quantify; we read it as consistent with meeting
the stretch target rather than definitively met.
This single-card measured $156.49$~s also came in faster than an earlier same-config 2-GPU
\emph{projection} of $\approx$$168.5$~s,\ 
because the single-card path avoids the cross-device split overhead that the 2-GPU
configuration carries.\ 
Second, it fits: the single-card fake-quant path peaks at $26.7$~GB (a peak measured on
an A100, used here to \emph{infer} the OOM on a $24$~GB 3090, since peak VRAM is not perfectly
card-invariant),\ 
while the fused build's $23.40$~GB is the direct 3090 measurement.\ 
Realizing the compute is therefore also what makes 1024px single-GPU feasible at all. INT8
moves from the slowest variant to the fastest.

\begin{table}[t]
\centering
\caption{1024px generation on the RTX~3090, $48$ steps, identical tuples. FP8/NF4/fake-quant
INT8 figures are published prior-work baselines on the same protocol~\cite{2606.12280}; the fused INT8 row is
this work's single-card measurement. ``GPUs'' is the number of cards required.}
\label{tab:1024}
\begin{tabular}{lrrr}
\toprule
Variant & s/image & GPUs & Note \\
\midrule
FP8 (prior work)            & $172.9$  & $2$ & dequant-to-bf16 on Ampere \\ 
NF4 (prior work)            & $164.5$  & $1$ & \\ 
INT8 fake-quant (prior work)& $184$--$185$ & $2$ & quantize then dequant to bf16 \\ 
\midrule
\textbf{Fused INT8 (ours)}  & $\mathbf{156.49}$ & $\mathbf{1}$ & peak $23.40$~GB, $4/4$ images \\ 
\bottomrule
\end{tabular}
\end{table}

\paragraph{Hardware specificity.}
The win is not universal, and we measured exactly where it disappears. Table~\ref{tab:hw}
runs the same fused kernel against the same fake-quant baseline at 1024px on an A100 and a
B200. On both the fused kernel is \emph{slower}: $1.38\times$ slower on the A100 and
$3.49\times$ slower on the B200.\ 
This is consistent with these cards' much faster native paths: the A100 delivers
$\approx 216$~TFLOPS bf16 and the B200 $\approx 587$--$610$~TFLOPS bf16,\ 
so the bf16 ``fake-quant'' matmul these cards run is already extremely fast, and the integer
path, competitive against the RTX~3090's much slower bf16, has nothing to win back. We
note one confound: the kernel is autotuned for the RTX~3090 (sm\_86) and is not retuned for
the A100 or B200, so kernel sub-optimality on those cards and their fast native bf16/FP8 paths
are not separable here. The deployment-map conclusion, use the fused kernel where there is
no fast native low-precision matmul, holds regardless of which factor dominates.
Quality is unaffected: at 1024px on the A100 the fused build scores PickScore $17.49$ against
the fake-quant build's $17.54$,\ 
confirming the kernel introduces no quality regression even where it is not the speed choice.

\begin{table}[t]
\centering
\caption{Hardware-specificity check at 1024px: fused INT8 vs.\ the fake-quant bf16 path on
datacenter Ampere (A100) and Blackwell (B200). The fused kernel loses on both, because
native bf16 on these cards is already fast.}
\label{tab:hw}
\begin{tabular}{lrrr}
\toprule
GPU & Fused (s/img) & Fake-quant (s/img) & Outcome \\
\midrule
A100 & $100.85$ & $73.04$ & fused $1.38\times$ slower \\ 
B200 & $98.72$  & $28.33$ & fused $3.49\times$ slower \\ 
\bottomrule
\end{tabular}
\end{table}

\paragraph{Ablations: the two obvious levers are dead ends.}
Two extensions suggest themselves for closing the gap between the per-GEMM and end-to-end
speedups, and we report both as honest negatives. Applying \texttt{torch.compile} to the
fused build at 768px yields $97.74$~s/image (effectively zero gain) because the compiler
graph-breaks at the dynamo-disabled linear layers.\ 
Replacing the bf16 attention with SageAttention's INT8 attention is blocked outright: the
DiT's head dimension of $256$ exceeds the library's $128$ limit, forcing a $100\%$ FP16
fallback and again zero gain ($97.82$~s/image).\ 
Neither lever is available without further engineering, which we scope as future work.

\begin{table}[t]
\centering
\caption{Ablations on the fused INT8 build at 768px. Both candidate extensions are inert on
this model.}
\label{tab:abl}
\begin{tabular}{lrl}
\toprule
Configuration & s/image & Why no gain \\
\midrule
Fused INT8 (baseline)    & $97.79$ & --- \\ 
$+$ \texttt{torch.compile} & $97.74$ & graph-breaks at disabled linears \\ 
$+$ SageAttention INT8     & $97.82$ & head\_dim $256 > 128$; FP16 fallback \\ 
\bottomrule
\end{tabular}
\end{table}

\begin{figure}[t]
\centering
\includegraphics[width=0.66\linewidth]{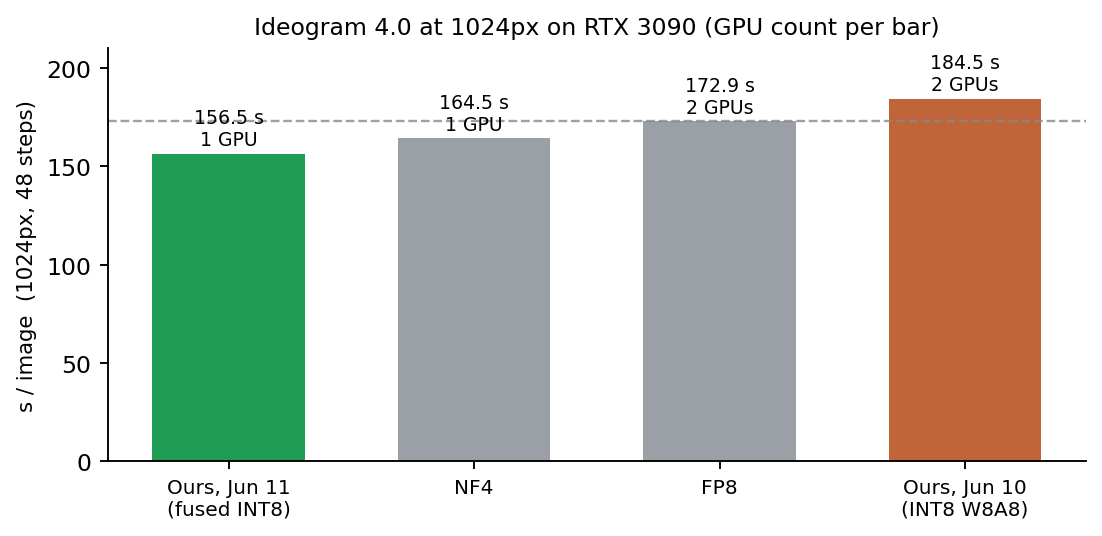}
\caption{End-to-end inference at 1024px on RTX 3090 hardware; the label on each bar gives the
number of cards that variant requires. Our fused INT8 build (June 11) runs on a single 3090 and
is faster than the single-card NF4 variant and the two-GPU FP8 variant; unlike FP8 it fits on one
card. Our earlier INT8 W8A8 build (June 10), also two-GPU, is the slowest~\cite{2606.12280}. 
INT8 moves from the slowest variant to the fastest.}
\label{fig:e2e}
\end{figure}

\section{Discussion and Limitations}
\label{sec:discussion}

\paragraph{The honest ceiling.}
The fused kernel is a large per-GEMM win (about $3.5\times$), yet the end-to-end gain is
$\approx$$9.5\%$ because the DiT linear GEMMs are only $\approx 12\%$ of the forward
(Results). This is not a shortcoming of the kernel but a structural fact about where the time
goes in this model: the remaining $\approx 88\%$ sits in attention, normalization, sampling
overhead, and the like, none of which the kernel touches. The two natural ways to extend the
win, compiling the whole forward into a fused graph and quantizing attention, are the
dead ends documented in our ablations:
\texttt{torch.compile} graph-breaks at the custom linears,\ 
and INT8 attention is blocked by the model's $256$ head dimension.\ 
We are explicit that absent progress on those two fronts, $\sim$$9.5\%$ is close to the
ceiling for this approach on this model.

\paragraph{Hardware specificity as deployment guidance.}
The most important caveat is also the cleanest result. The kernel is a win on consumer
Ampere precisely because consumer Ampere has no fast FP8 or bf16 alternative; on the A100 and
B200, whose native bf16/FP8 paths run at $216$ and $587$--$610$~TFLOPS,\ 
the fused kernel loses by $1.38\times$ and $3.49\times$ respectively.\ 
We frame this as precise guidance rather than a limitation: deploy the fused INT8 kernel when,
and only when, the target lacks a fast native low-precision matmul: the consumer-Ampere
regime. On datacenter or Blackwell hardware, run the native bf16/FP8 path.

\paragraph{The memory result.}
Realizing the compute also changes the memory envelope in a way that matters operationally.
At 1024px the fused build peaks at $23.40$~GB and fits a single $24$~GB RTX~3090 (a direct
3090 measurement),\ 
while the fake-quant path peaks at $26.7$~GB (measured on an A100, from which we infer the
OOM on a $24$~GB 3090, peak VRAM not being perfectly card-invariant).\ 
The speed win and the single-card feasibility are therefore two aspects of the same fix.

\subsection{Threats to Validity}

\paragraph{Internal.}
End-to-end s/image is measured on a $4$-prompt subset.\ 
We argue this is benign because at fixed steps and resolution latency is nearly
prompt-independent, but $n=4$ is small and we report it as such. The headline 1024px
comparison mixes our single-card fused measurement against FP8/INT8 baselines that were
recorded on two GPUs~\cite{2606.12280};\ 
those two-GPU numbers carry a multi-GPU split overhead that a single-card run does not, so the
fused-vs-FP8 margin should be read with that asymmetry in mind (the NF4 comparison is
single-card on both sides and is the cleanest apples-to-apples). We also did not re-measure
OCR/text-rendering NED at 1024px for the fused build; quality parity at 1024px is established
on PickScore only (A100),\ 
and the typography metric, the most quantization-fragile one, is carried over from the
recipe rather than re-verified for the fused path at any resolution, which we flag as an open
check. Two further items are left to future work. First, the $\approx 12\%$ GEMM share is
\emph{implied by} the measured per-GEMM and end-to-end speedups via Amdahl's law, not an
independent forward-pass profile; a direct layer-wise profile to confirm the GEMM fraction
remains to be done. Second, our unfused baseline is the standard PyTorch
\texttt{torch.\_int\_mm} two-step path; a tuned-library INT8 GEMM baseline
(cuBLASLt / CUTLASS) was planned but is not reported, and beating a tuned vendor kernel is
left to future work.

\paragraph{External.}
All results are specific to one model (Ideogram~4.0) on consumer Ampere; we have shown directly
that they do not transfer to A100 or B200.\ 
We make no claim about other diffusion transformers, other quantization recipes, or other
GPU generations.

\paragraph{Construct.}
PickScore and CLIPScore are proxies for human-perceived image quality and prompt adherence,
not the quantity of interest itself. They are the field-standard proxies and we use them for
comparability, but a parity claim on these metrics is a parity claim on the proxies.

\section{Availability}

The fused INT8 build of Ideogram~4.0 is the gated, private checkpoint repository
\texttt{transformerlab/\allowbreak ideogram-4-\allowbreak int8-fused} under the
\texttt{transformerlab} organization.
It is distributed under the Ideogram~4.0 non-commercial, research-only license, inherited from
the base model; access is gated and the weights may be used for research purposes only.

\section{Conclusion and Future Work}

Production post-training INT8 for diffusion transformers on consumer GPUs is, in the form most
commonly deployed, fake-quant: it dequantizes to bf16 and never touches the integer tensor
cores, which is why ``INT8'' can be the slowest variant on a card that has fast native INT8.
A single fused Triton INT8 GEMM that keeps the multiply on the integer tensor cores and folds
per-token$\times$per-channel dequantization and bias into the epilogue removes this artifact:
its int8$\times$int8$\rightarrow$int32 accumulation is bit-exact against \texttt{torch.\_int\_mm}
(cosine $1.0$, no NaNs),\ 
it is $2.8$--$4.2\times$ faster per GEMM, $\approx$$9$--$10\%$ faster end to end at
768px, and at 1024px it runs on a single RTX~3090 at $156.5$~s/image, faster than both FP8
and NF4, turning INT8 from the slowest variant into the fastest and making 1024px
single-GPU feasible. The win is specific to consumer Ampere, and we have mapped exactly where
it does not apply.

Three next steps follow directly from the limitations. First, a graph-safe fused custom op
that survives \texttt{torch.compile} without graph-breaking, to compose the GEMM win with
whole-forward compilation. Second, an INT8 attention kernel that supports a head dimension of
$256$, since the existing library is blocked there and attention dominates the non-GEMM
budget. Third, a direct two-GPU measurement at 1024px to retire the single-vs-two-GPU
asymmetry in the baseline comparison. Together these target the $\approx 88\%$ of the forward
that the present kernel does not yet touch.

\bibliographystyle{unsrt}  
\bibliography{references}

\end{document}